# AutoScore-Imbalance: An Interpretable Machine Learning Tool for Development of Clinical Scores with Rare Events Data

Han Yuan[1], Feng Xie[1], Marcus Eng Hock Ong[1,2,3], Yilin Ning[1], Marcel Lucas Chee[4], Seyed Ehsan Saffari[1], Hairil Rizal Abdullah[1,5], Benjamin Alan Goldstein[1,8], Bibhas Chakraborty[1,8,9], Nan Liu[1,3,10]*

[1] Duke-NUS Medical School, National University of Singapore, Singapore, Singapore
[2] Department of Emergency Medicine, Singapore General Hospital, Singapore, Singapore
[3] Health Services Research Centre, Singapore Health Services, Singapore
[4] Faculty of Medicine, Nursing and Health Sciences, Monash University, Melbourne, Australia
[5] Department of Anaesthesiology, Singapore General Hospital, Singapore, Singapore
[8] Department of Biostatistics and Bioinformatics, Duke University, Durham, NC, United States
[9] Department of Statistics and Data Science, National University of Singapore, Singapore, Singapore
[10] Institute of Data Science, National University of Singapore, Singapore, Singapore

* Corresponding Author
Nan Liu
Programme in Health Services and Systems Research
Duke-NUS Medical School
8 College Road
Singapore 169857
Singapore
Phone: +65 6601 6503
Email: liu.nan@duke-nus.edu.sg

## Abstract

**Background**: Medical decision-making impacts both individual and public health. Clinical scores are commonly used among a wide variety of decision-making models for determining the degree of disease deterioration at the bedside. AutoScore was proposed as a useful clinical score generator based on machine learning and a generalized linear model. Its current framework, however, still leaves room for improvement when addressing unbalanced data of rare events.

**Objective**: This paper aims to propose and validate AutoScore-Imbalance, which extends the original AutoScore to generate accurate clinical scores on unbalanced datasets.

**Methods**: Using machine intelligence approaches, we developed AutoScore-Imbalance, which comprises three components: training dataset optimization, sample weight optimization, and adjusted AutoScore. Baseline techniques for performance comparison included the original AutoScore, full logistic regression, stepwise logistic regression, least absolute shrinkage and selection operator (LASSO), full random forest, and random forest with reduced number of features. The models were evaluated on the basis of their area under the curve (AUC) in the receiver operating characteristic analysis and balanced accuracy (i.e., mean value of sensitivity and specificity). By utilizing a publicly accessible dataset from Beth Israel Deaconess Medical Center, we assessed the proposed model and baseline approaches in the prediction of inpatient mortality.

**Results**: AutoScore-Imbalance outperformed baselines in terms of AUC and balanced accuracy. The nine-variable AutoScore-Imbalance sub-model achieved the highest AUC of 0.786 (0.732-0.839) while the eleven-variable original AutoScore obtained an AUC of 0.723 (0.663-0.783), and the logistic regression with 21 variables obtained an AUC of 0.743 (0.685-0.800). The AutoScore-Imbalance sub-model (using down-sampling algorithm) yielded an AUC of 0. 0.771 (0.718-0.823) with only five variables, demonstrating a good balance between performance and variable sparsity. Furthermore, AutoScore-Imbalance obtained the highest balanced accuracy of 0.757 (0.702-0.805), compared to 0.698 (0.643-0.753) by the original AutoScore and the maximum of 0.720 (0.664-0.769) by other baseline models).

**Conclusions**: We have developed an interpretable tool to handle clinical data imbalance, presented its structure, and demonstrated its superiority over baselines. The AutoScore-Imbalance tool has the potential to be applied to highly unbalanced datasets to gain further insight into rare medical events and to facilitate real-world clinical decision-making.

# 1 Introduction

In the field of medicine, decision-making encompasses diagnosis, treatment, disease prediction, and everyday conditions that impact individual and public health [1]. The increasing collection of clinical data, such as the growing electronic health records (EHRs) in hospitals and advances in automated machine learning (AutoML), have facilitated automatic medical decision-making [2, 3]. In general, clinicians prefer transparent and interpretable "glass box" models to complex "black box" models, such as artificial neural networks (ANN) [4, 5]. Interpretable models can be explained or presented in an understandable manner to a human being [6]. SHapley Additive exPlanations (SHAP) is an innovative approach to providing interpretability to "black box" models [7]. Likewise, local interpretable model-agnostic explanations (LIME) is another technique that explains classifiers' predictions based on learning interpretable models around "black box" predictions [8]. However, both SHAP and LIME can be viewed as post hoc explanation methods that are not transparent enough for clinicians, who are more inclined to inherently transparent models like logistic regression (LR) [9].

Therefore, generic clinical scores for a variety of applications are widely accepted and used by clinicians and nurses in hospitals [10-12]. Such scores take advantage of integer score points and categorized predictors to identify clinical outcomes, trigger better care, and improve prognoses [13]. Clinical scores are often derived from expert consensus or through cohort analyses using traditional methods such as logistic regression [14, 15].

To aid the development and validation of interpretable clinical scores, Xie, et al. [16] proposed AutoScore, a novel scoring framework for AutoML. Despite AutoScore's ability to generate clinical score systems for a wide range of medical applications [17, 18], it may not perform well when datasets are unbalanced due to low prevalence of outcomes, i.e., rare medical events. Unbalanced datasets often make predictive models unreliable since they tend to focus on the dominant class and disregard the rare one [19]. Consequently, data imbalance may lead to poor prediction capabilities [20-22].

Data imbalance has been addressed through a variety of approaches. Researchers traditionally used up-sampling (over-sampling) of minority samples, down-sampling (under-sampling) of majority samples, and sample weight adjustment to ensure a balanced distribution of classes [23]. The synthetic minority over-sampling technique (SMOTE) is a popular algorithm dealing with unbalanced datasets, which synthesizes new minority samples from several closest neighbors of the real minority samples [24, 25]. In medical domain, the unbalanced nature of many datasets has prompted researchers to propose novel approaches. Rahman et al. developed cluster-based under-sampling technique [26], while Khalilia et al. sampled data to subgroups to build multiple random forest models (RF) and then ensembled these models to deal with data imbalance [27]. Li et al. introduced Gaussian type fuzzy membership function for data down-sampling [28]. In recent years, generative adversarial networks (GANs) [29] have been widely used in medical image syntheses to create new images for model development. The GAN technique can also be applied to the generation of structured data, which is useful for augmenting the proportion of minority samples in datasets and even improving the sample category distribution [30].

In this study, we sought to integrate methodologies for handling data imbalance into AutoScore and create an automated framework that allows for reliable risk scores to be derived even from unbalanced datasets.

## 2 Methods
### 2.1 The AutoScore Framework
AutoScore [16] is a machine learning-based clinical score generator, consisting of six modules. Module 1 uses a random forest to rank variables according to their importance. Module 2 transforms variables by categorizing continuous variables to improve interpretation and cope with nonlinearity. Module 3 assigns scores to each variable based on a logistic regression model. Depending on the trade-off between model complexity and predictive performance, Module 4 determines the number of variables to include in the scoring model. In Module 5, where clinical knowledge is incorporated, cutoff points can be fine-tuned when categorizing continuous variables. Module 6 evaluates the performance of the score in a separate test dataset. The AutoScore framework provides a systematic and automated approach to quick development of a scoring system, combining both the advantage of machine learning in the ability to discriminate and the strength of point-based scores in its interpretability.

**2.2 Proposed AutoScore-Imbalance Framework**

To deal with data imbalance and automate the development of sparse clinical scores, we proposed AutoScore-Imbalance, a novel extension to the original AutoScore framework. AutoScore-Imbalance adopts a nested structure to combine and reorganize AutoScore and its individual modules. It is comprised of three blocks including Block A (training data optimization), Block B (sample weights optimization), and Block C (final score derivation and evaluation). The AutoScore-Imbalance framework is illustrated in Figure 1. By using resampling and data synthesis techniques, Block A adjusts the raw unbalanced training dataset. Block B is designed to optimize sample (observation) weights, which are tuned to correct any imperfections that may lead to bias in the class proportion [31]. Block C adapts the original AutoScore workflow to derive risk scores using the relatively balanced datasets obtained in Block A and the sample weights acquired from Block B. Modules A and B are newly introduced for AutoScore-Imbalance and Modules 1-6 remain the same as those in the original AutoScore framework.

**2.2.1 Block A: Training Data Optimization**

Using the unbalanced training dataset as input, Block A manipulates the data to produce a reasonably balanced dataset. Variable quantity of scoring systems is used as a hyperparameter in Block A and Block B for intermediate evaluations. As with random forest, we set this hyperparameter as the square root of the total number of variables [32].

Similarly to AutoScore, AutoScore-Imbalance divides a full dataset into three parts: training data $D$, validation data, and test data. The training dataset $D$ is used to derive scores, the validation dataset is used for intermediate evaluation and parameter optimization, and the test dataset is reserved as unseen data for model performance assessment [26]. The data-balancing methods in Block A are only applied to training data. The training dataset $D$ of $N$ samples is defined as follows,

$$D = \{D_i\}, i = 1, 2, 3, \ldots, N-1, N \tag{1}$$

where $i$ represents the $i$th subject. $D$ contains $N_p$ minority samples (positive samples) and $N_n$ majority samples (negative samples). In this study, we assume that rare clinical events are positive outcomes. The minority prevalence rate $P$ is defined as $N_p/N$. The objective of Module A is to increase the minority rate from $P$ to $P'$ ($P < P' \leq 0.5$) in the manipulated training dataset $D'$. By using up-sampling or other data augmentation techniques, the corresponding number of minority samples $N_p'$ in the processed dataset is:

$$N_p' = round\left(\frac{N_n}{(1-P')} - N_n\right) \tag{2}$$

When we use down-sampling to reduce sample size, the number of majority samples $N_n'$ will become:

$$N_n' = round\left(\frac{N_p}{P'} - N_p\right) \tag{3}$$

In this study, we use integer $\alpha > 0$ to denote the up-sampling ratio, which corresponds to the quotient in a Euclidean division of $N_p'$ by $N_p$, i.e.,

$$N_p' = \alpha N_p + r, \tag{4}$$

where $r$ is the integer remainder with $0 \leq r < N_p$. In up-sampling operation, $\alpha$ stands for the replication times of $N_p$, and $r$ is the number of additional samples drawn from $N_p$ data. For example, if we increase the minority sample size from $N_p$ (100) to $N_p'$ (205), $\alpha$ is 2 and $r$ is 5. The up-sampled dataset by Block A will include $2 * N_p$ minority samples (the original $N_p$ and the duplicated $N_p$ samples), 5 randomly selected minority samples from the original $N_p$ data, and $N_n$ majority samples. In SMOTE-based data augmentation, the number of synthesized samples is $(\alpha - 1) * N_p + r$, which is 105 based on the former example. Since SMOTE cannot generate non-integer times of artificial data, we achieve data augmentation in two steps: first, we use SMOTE to create $(\alpha - 1) * N_p$ (i.e., 100) synthetic samples from the original $N_p$ minority data; second, we randomly select five samples from the original $N_p$ samples, and then apply SMOTE to produce five synthetic ones from them.

There are three types of methods for training data optimization in Module A: resampling methods, data synthesis methods, and hybrid methods that combine both resampling and data synthesis strategies. Resampling methods include up-sampling (US) of minority samples and down-sampling (DS) of majority samples. Data synthesis methods include SMOTE and GAN. These methods and their hybrid versions are used in Module A to generate a variety of processed datasets with different minority rates $P'$ ($P < P' \leq 0.5$). Subsequently, each processed dataset goes through Modules 1, 2, and 3 to construct risk scores using different number of top-ranking variables. These scores are evaluated based on the area under the curve (AUC) in the receiver operating characteristic (ROC) analysis applied to the validation dataset, and the optimal processed dataset will be returned. Next, we describe in detail the various methods included in Module A, as shown in Table 1.

**Resampling Methods:** Simple data resampling techniques include up-sampling of minority samples and down-sampling of majority samples. They are the most commonly used techniques for dealing with data imbalance [33, 34].

**Data Synthesis Methods:** SMOTE creates realistic "pseudo" minority samples through the following steps [35]: (i) Select $Z$ nearest neighbors of each minority sample; (ii) Calculate differences between the sample and its $Z$ nearest neighbors; (iii) Multiply differences by random numbers $L$ ($0 < L < 1$); (iv) Add those products to the sample to generate a synthetic sample. By default, SMOTE creates $Z$ ($Z$ is an integer) sets of minority samples, i.e., $Z * N_p$ samples, but cannot directly synthesize $Z * N_p + C$ ($0 < C < N_p$) artificial samples. Therefore, we customized the original SMOTE DMwR package [36] to fit our needs. GAN was initially proposed by Goodfellow et al. to generate synthetic images [29] and Xu et al. extended its application to structured data generation [37]. GAN has two adversarial components: a generative model and a discriminative model. Through iterative learning, the pseudo data generated by the generative model becomes increasingly similar to the real data. In this study, we use GAN to generate synthetic minority samples in the training dataset.

**Hybrid Methods:** Additionally, we explore several hybrid techniques that combine the resampling and data synthesis methods. We first increase the minority sample quantity to an intermediate level by up-sampling, SMOTE, or GAN. Next, we reduce the sample size from the majority class to a specified level through down-sampling. Table 1 shows three hybrid methods used in our demonstration: up-sampling + down-sampling, SMOTE + down-sampling, and GAN + down-sampling.

### 2.2.2 Block B: Sample Weights Optimization

Block B is designed to derive optimal sample weights for the majority and minority samples generated from Block A. The sample weight is defined as the contribution of each subject $D_i$ to the loss function, with minority samples assigned a weightage greater than one to make it more costly to erroneous predictions. With the sample weights, our proposed model no longer solely concentrates on the majority samples and ignores the minority samples [38]. The sample weight of the majority sample is always set as one. Block B receives the optimal dataset from Block A, and outputs the optimal sample weights for samples in the optimal, processed training dataset. It selects the optimal sample weights for minority samples in a grid search ranging from 1 to $w_{max} = N'_n/N'_p$ (to be rounded up to an integer) with a pre-set integer step, $s$, using the AUC on the validation dataset as the criterion [31].

### 2.2.3 Block C: Final Score Derivation and Evaluation

Using the relatively balanced training dataset obtained from Block A and the optimal sample weights obtained from Block B as the inputs, Block C employs the original AutoScore framework (but beginning from Module 2) to generate sparse clinical scores. Module 2 converts continuous variables in the relatively balanced, restructured training dataset into categorical variables. In contrast to the original Module 3 in AutoScore, which uses an unweighted logistic regression model, the Module 3 in Block C applies a weighted logistic regression model to the processed training dataset using the sample weights obtained from Block B. Using a parsimony plot, Module 4 determines the number of variables to include in the clinical score using a parsimony plot (top-ranked variables are chosen when there is no substantial improvement in AUC values with the addition of more variables). Module 5 allows users to customize cutoffs based on their domain knowledge. Module 6 evaluates the derived clinical score based on multiple performance evaluation metrics. Overall, Block C, which is the last step of the AutoScore-Imbalance framework, produces a standard clinical score table used for subsequent risk prediction.

### 2.3 Experiments

We used de-identified critical care unit data from 44,918 admission episodes (3,958 positive admissions) of the Beth Israel Deaconess Medical Center between 2001 and 2012 (MIMIC-III dataset) [24]. In academic research, it is generally considered that the majority-minority class ratio in high-class imbalance scenario should be less than or equal to 1% [39, 40]. In our study, we randomly selected 404 positive and 40,000 negative admission episodes to create a dataset with 1% positive rate for demonstrating our methods under highly imbalanced conditions [41]. To derive and evaluate the scoring models, we split the entire dataset into three parts: training dataset (60%), validation dataset (20%), and test dataset (20%).

We compared the scoring model derived by AutoScore-Imbalance with that of the original AutoScore, full logistic regression, stepwise logistic regression, LASSO, full random forest, and random forest with reduced number of variables. We chose the "optimal" thresholds as the points nearest to the upper-left corner in the ROC curves to calculate performance metrics. In addition to commonly used metrics (AUC, sensitivity, specificity, negative predictive value (NPV), positive predictive value (PPV)), we used balanced accuracy (i.e., the average of the sensitivity and specificity values) to evaluate various model predictions on unbalanced datasets [42]. Sensitivity, specificity, NPV, PPV, and balanced accuracy were calculated with each model's optimal threshold [43] and their corresponding 95% confidence intervals (CIs) were obtained via bootstrapping [44].

## 2.4 Code Availability

We implemented AutoScore-Imbalance in R based on the AutoScore package [45] and made corresponding software package available on GitHub (https://github.com/nliulab/AutoScore-Imbalance). In addition, we intend to incorporate AutoScore-Imbalance into the AutoScore main package.

## 3 Results

This study analyzed 40,404 admission episodes. Specifically, the training dataset consisted of 24,244 episodes (60%, containing 244 positive samples) while the validation dataset consisted of 8,080 episodes (20%, containing 80 positive samples) and the test dataset comprised 8,080 episodes (20%, containing 80 positive samples). We created a total of nine scoring models, including AutoScore. Figure 2 illustrates the parsimony plots for the original AutoScore and eight sub-models of the AutoScore-Imbalance framework. We manually selected the "near-optimal" number of variables to ensure that the performance could not be significantly improved by including additional variables in the models. When compared with the original AutoScore, AutoScore-Imbalance sub-models achieved the "near-optimal" solutions with fewer variables, except for its sub-model (GAN, $e = 500$), which was insufficiently trained due to a small number of epochs in deep learning frameworks. Furthermore, the results in Table 2 shows an enhanced performance of GAN-based scoring models with increased training epochs.

We summarize the performance of different scoring models in Table 2. The original AutoScore selected 11 variables and achieved an AUC of 0.723 (0.663-0.783). Despite using fewer variables (apart from GAN [$e = 500$]), all sub-models of AutoScore-Imbalance had higher AUC and balanced accuracy values than the original AutoScore. Particularly, AutoScore-Imbalance (down-sampling) achieved an AUC of 0.771 (0.718-0.823) with only five variables—less than half of the variables required by the original AutoScore. Furthermore, AutoScore-Imbalance yielded the highest balanced accuracy of 0.757 (0.702-0.805) using 10 predictors, as compared to 0.698 (0.643-0.753) in the original AutoScore with 11 predictors and the maximum of 0.720 (0.664-0.769) in other baseline models). In terms of both AUC and balanced accuracy, AutoScore-Imbalance sub-models (SMOTE, up-sampling, up-sampling + down-sampling) outperformed all baseline methods.

Table 3 shows the selected variables by the original AutoScore, and eight sub-models of the AutoScore-Imbalance framework. As shown in the table, blood urea nitrogen was selected by all AutoScore-Imbalance scoring models, but not by the original AutoScore. Several variables such as heart rate, age, lactate, and respiration rate were selected by the original AutoScore and most of AutoScore-Imbalance sub-models.

With AutoScore-Imbalance, a scoring table can be generated for direct application to clinical practice. This score ranging from 0 to 100 (the range could be adjusted based on clinical needs) is used to identify patients at risk of suffering from adverse events. The minimum score is 0 which stands for no risk while the maximum score of 100 means the highest risk. As an example, Table 4 presents a summary of fine-tuned scoring tables for inpatient mortality prediction based on the nine-variable AutoScore-Imbalance sub-model (US+DS), five-variable AutoScore-Imbalance sub-model (DS), and eleven-variable original AutoScore scoring model.

# 4 Discussion

In this study, we developed an interpretable machine learning model that coped with data imbalance and generated trustworthy clinical scores. Using an unbalanced real-world dataset, the proposed AutoScore-Imbalance framework was found to achieve an improved prediction performance than the original AutoScore with fewer variables. Due to its sparsity and parsimony in variable selection, a clinical score derived from AutoScore-Imbalance is practical for use at the bedside [46].

When studying various approaches to handle data imbalance, conventional methods (up-sampling, down-sampling, and SMOTE) have proven to be effective in handling unbalanced datasets for the development of risk scores. Additionally, modern technique like GAN displayed comparable performance and outperformed the original AutoScore when training epochs were increased. GAN needs sufficient training epochs to generate reliable synthetic samples [47, 48]. Therefore, a suitable training epoch for GAN should be determined upon assessment of a validation dataset.

Compared with AutoScore and AutoScore-Imbalance, baseline methods have an intrinsic advantage, as they can build scoring models using high-resolution, continuous variables rather than categorized variables [49]. This superiority, however, poses a limitation to the utility of baseline models in clinical risk prediction, where sparse and itemized scores are favored. In this regard, it is noteworthy that AutoScore-Imbalance produced improved prediction results while preserving score sparsity and its clinical usability [50]. Furthermore, among all baseline models, LASSO demonstrated the best prediction ability, which is in accordance with the fact that LASSO is effective in identifying important variables in unbalanced datasets [51, 52].

The strength of AutoScore-Imbalance lies in its ability to address data imbalance while at the same time producing reliable and sparse clinical scores. The outputs of AutoScore-Imbalance—integer-based, itemized clinical scores—are more preferred by clinicians than complicated "black box" models [53]. In this study, we demonstrated the effectiveness of proposed AutoScore-Imbalance framework using real-world data obtained from MIMIC-III database. This novel framework provides a practical and referable two-step pipeline for modifying training data and adjusting sample weights to handle data imbalance, which is not specifically limited to clinical applications.

Moreover, AutoScore-Imbalance is designed in a modular manner to make it easy to incorporate other state-of-the-art techniques for efficient score derivation and evaluation.

Our study has limitations. First, we evaluated AutoScore-Imbalance using only one minority sample ratio (1%); thus, we intend to test other minority sample ratios in future studies. Second, for the purposes of demonstration, we examined only one clinical application using the MIMIC III database. Moreover, given the low percentage of positive samples, there were only 80 positive cases in the test set, which led to relatively low PPV values and overlapping CIs of evaluation metrics for all methods. It is therefore necessary to perform additional validations under different settings. Third, the heterogeneity of clinical applications prevented us from recommending the best method to handle data imbalance. In general, SMOTE is a popular tool for augmenting data, but might not be effective in certain scenarios, as with other sampling techniques [54]. And, it is worth considering GAN for dealing with high-dimensional data [55]. Lastly, our method was designed for tabular data without considering time series data [56, 57]. To develop a complete AutoScore-Imbalance solution, further studies will be required to extend its application to longitudinal data [58].

## 5 Conclusion

We proposed an interpretable machine learning-based AutoScore-Imbalance framework for automatic clinical score generation that addresses data imbalance. Compared with baseline models, this innovative framework presented a capability of developing good-performing and reliable, yet interpretable clinical scores on unbalanced datasets. We anticipate that this score generator will hold great potential in creating and evaluating sparse and itemized clinical scores in a variety of settings.

**Data Availability**

We used de-identified critical care unit data from the Beth Israel Deaconess Medical Center between 2001 and 2012 (MIMIC-III dataset) [24], which is available at https://archive.physionet.org/physiobank/database/mimic3cdb/.

**Author Contributions**

NL conceived and supervised the study. HY and NL developed the AutoScore-Imbalance algorithm and wrote the first draft of the manuscript. HY analyzed the data. HY, FX, MO,

**Figure Legends**

Figure 1. Flowchart of the AutoScore-Imbalance framework

Figure 2: Parsimony plots of the original AutoScore and AutoScore-Imbalance sub-models on validation datasets (the orange diamond indicates the number of variables selected for each model)

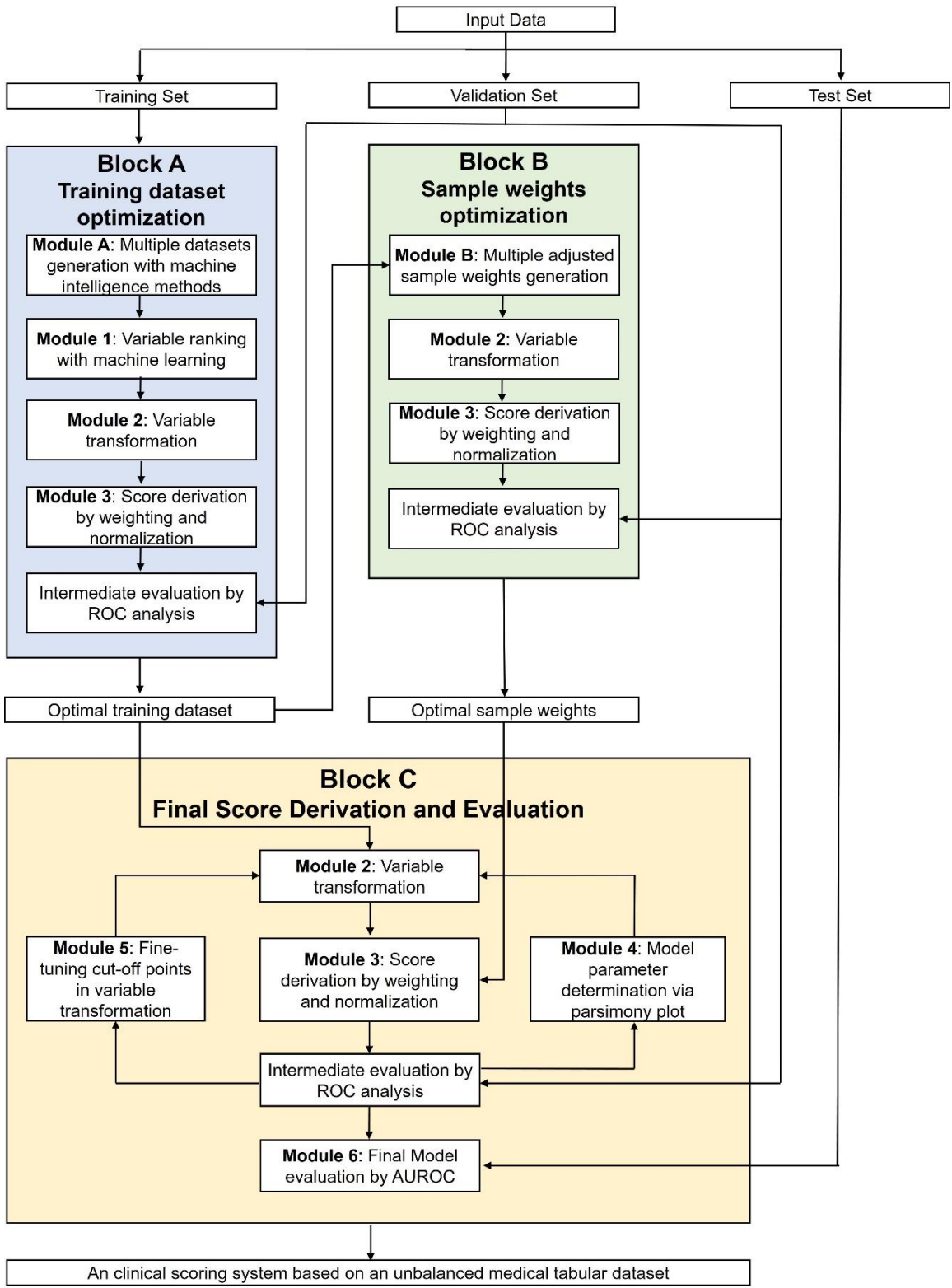

Figure 1. Flowchart of the AutoScore-Imbalance framework

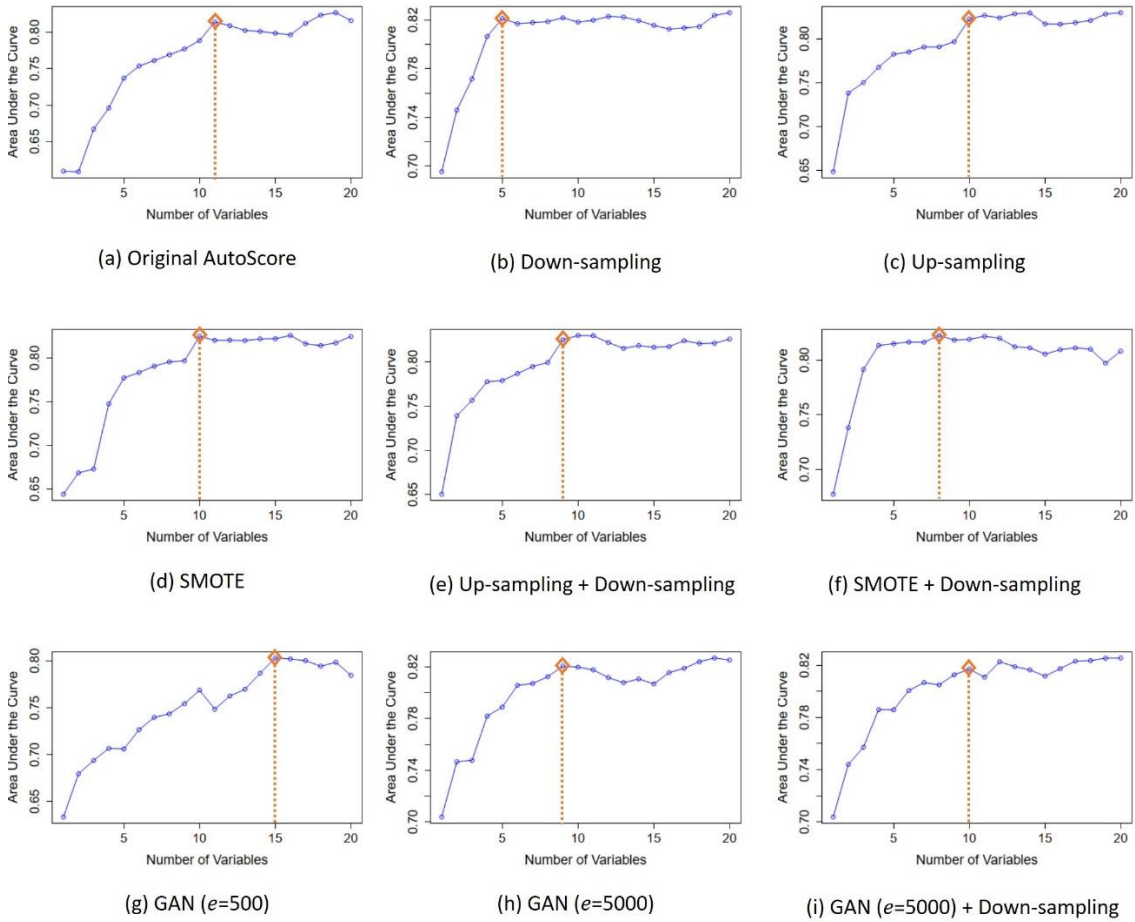

Figure 2: Parsimony plots of the original AutoScore and AutoScore-Imbalance sub-models on validation datasets (the orange diamond indicates the number of variables selected for each model)

Table 1: List of methods and their detailed algorithms used in Block A (training data optimization)

| Type | Method | Algorithm |
|---|---|---|
| Resampling Methods | Down-sampling | $D' = \{D'_1, D'_2\}$, where $D'_1$ is $D_p$, and $D'_2$ is a set of $N'_n$ selected samples without replacement from $N_n$ |
| | Up-sampling | $D' = \{D'_1, D'_2, D'_3\}$, where $D'_1$ is $D_n$, $D'_2$ is $\alpha$ times replication of $D_p$, and $D'_3$ is $r$ selected samples without replacement from $N_p$ |
| Data Synthesis Methods | SMOTE | $D' = \{D'_1, D'_2, D'_3\}$, where $D'_1$ is $D$, $D'_2$ is a set of $(\alpha - 1) * N_p$ synthetic samples obtained from $N_p$ through SMOTE, and $D'_3$ is a set of $r$ synthetic samples obtained from $r$ randomly selected samples of $N_p$ through SMOTE |
| | GAN | $D' = \{D'_1, D'_2\}$, where $D'_1$ is $D$, and $D'_2$ is a set of $(N'_p - N_p)$ synthetic minority samples obtained from $D$ through GAN |
| Hybrid Methods | Up-sampling + Down-sampling | $D'_h$ is an intermediate dataset with minority rate of $(P + P')/2$ generated from $D$ through up-sampling, and the final dataset $D'$ with minority rate of $P'$ is created from $D'_h$ through down-sampling |
| | SMOTE + Down-sampling | $D'_h$ is an intermediate dataset with minority rate of $(P + P')/2$ generated from $D$ through SMOTE, and the final dataset $D'$ with minority rate of $P'$ is created from $D'_h$ through down-sampling |
| | GAN + Down-sampling | $D'_h$ is an intermediate dataset with minority rate of $(P + P')/2$ generated from $D$ through GAN, and the final dataset $D'$ with minority rate of $P'$ is created from $D'_h$ through down-sampling |

$D$: The original training dataset

$D'$: The training datasets after Module A processing

$D_p$: The minority samples in $D$

$D_n$: The majority samples in $D$

$D'_1, D'_2, D'_3$: The first, second, and third part in $D'$, individually

$D'_h$: The intermediate dataset in hybrid methods

$N$: The total sample size in $D$

$N_p$: The minority sample size in $D$

$N_n$: The majority sample size in $D$

$N'_p$: The minority sample size in $D'$ (See Equation (2))

$N'_n$: The majority sample size in $D'$ (See Equation (3))

$P$: The minority rate in $D$

$P'$: The minority rate in $D'$

α: the quotient in the Euclidean division of $N'_p$ by $N_p$ (See Equation (4))

$r$: The remainder part of the Euclidean division of $N'_p$ by $N_p$ (See Equation (4))

Table 2: Performance of the original AutoScore, AutoScore-Imbalance, and baselines

| Models | | $m$ [a] | Threshold [b] | AUC [c] | Sensitivity [d] | Specificity [e] | Balanced Accuracy [f] | NPV [g] | PPV [h] |
|---|---|---|---|---|---|---|---|---|---|
| AutoScore | | 11 | 58 | 0.723 (0.663-0.783) | 0.700 (0.600-0.800) | 0.696 (0.686-0.706) | 0.698 (0.643-0.753) | 0.996 (0.994-0.997) | 0.022 (0.019-0.026) |
| Full LR | | 21 | 0.00736 | 0.743 (0.685-0.801) | 0.787 (0.700-0.875) | 0.602 (0.591-0.612) | 0.695 (0.646-0.744) | 0.996 (0.995-0.998) | 0.019 (0.017-0.022) |
| Stepwise LR | | 16 | 0.0108 | 0.737 (0.679-0.796) | 0.637 (0.537-0.750) | 0.748 (0.738-0.757) | 0.693 (0.638-0.754) | 0.995 (0.994-0.997) | 0.025 (0.021-0.029) |
| LASSO | | 6 | -4.586 | 0.768 (0.716-0.820) | 0.738 (0.637-0.825) | 0.702 (0.691-0.712) | 0.720 (0.664-0.769) | 0.996 (0.995-0.998) | 0.024 (0.021-0.027) |
| Full RF | | 21 | 0.005 | 0.743 (0.685-0.800) | 0.775 (0.675-0.863) | 0.602 (0.591-0.613) | 0.689 (0.633-0.738) | 0.996 (0.995-0.998) | 0.019 (0.017-0.021) |
| Parsimony RF | | 11 | 0.005 | 0.714 (0.655-0.772) | 0.750 (0.650-0.838) | 0.592 (0.581-0.602) | 0.671 (0.616-0.720) | 0.996 (0.994-0.997) | 0.018 (0.016-0.020) |
| AutoScore-Imbalance | SMOTE | 10 | 55 | 0.779 (0.727-0.832) | 0.775 (0.675-0.863) | 0.685 (0.675-0.695) | 0.730 (0.675-0.779) | 0.997 (0.995-0.998) | 0.024 (0.021-0.027) |
| | US | 10 | 57 | 0.780 (0.724-0.835) | 0.738 (0.637-0.825) | 0.776 (0.767-0.785) | 0.757 (0.702-0.805) | 0.997 (0.995-0.998) | 0.032 (0.028-0.036) |
| | DS | 5 | 68 | 0.771 (0.718-0.823) | 0.537 (0.437-0.637) | 0.873 (0.865-0.880) | 0.705 (0.651-0.759) | 0.995 (0.994-0.996) | 0.041 (0.032-0.048) |
| | US + DS | 9 | 55 | 0.786 (0.732-0.839) | 0.725 (0.625-0.825) | 0.758 (0.749-0.768) | 0.742 (0.687-0.797) | 0.996 (0.995-0.998) | 0.029 (0.025-0.033) |
| | SMOTE + DS | 8 | 55 | 0.767 (0.714-0.82) | 0.750 (0.650-0.838) | 0.689 (0.678-0.699) | 0.720 (0.664-0.769) | 0.996 (0.995-0.998) | 0.024 (0.020-0.026) |
| | GAN ($e = 500$) | 15 | 49 | 0.744 (0.686-0.802) | 0.625 (0.525-0.725) | 0.795 (0.786-0.803) | 0.710 (0.656-0.764) | 0.995 (0.994-0.997) | 0.03 (0.024-0.035) |
| | GAN ($e = 5000$) | 9 | 36 | 0.753 (0.704-0.803) | 0.725 (0.625-0.825) | 0.690 (0.679-0.700) | 0.708 (0.652-0.763) | 0.996 (0.995-0.997) | 0.023 (0.020-0.026) |
| | GAN ($e = 5000$) + DS | 10 | 36 | 0.759 (0.710-0.809) | 0.738 (0.637-0.838) | 0.699 (0.689-0.709) | 0.719 (0.663-0.774) | 0.996 (0.995-0.998) | 0.024 (0.021-0.027) |

[a] The number of variables included in each model.

[b] Optimal cutoff values, defined as the points nearest to the upper-left corner in the ROC curves.

[c] AUC: the area under the ROC curve.

[d] Sensitivity = TP / (TP + FN), TP: true positive, FN: false negative.

[e] Specificity = TN / (TN + FP), TN: true negative, FP: false positive.

[f] Balanced Accuracy = (Sensitivity + Specificity)/2.

[g] NPV: negative predictive value = TN/ (TN + FN)

[h] PPV: positive predictive value = TP / (TP + FP)
LR: Logistic regression.
LASSO: Least absolute shrinkage and selection operator.
RF: Random forest.
SMOTE: Synthetic minority over-sampling technique.
US: Up-sampling.
DS: Down-sampling.
GAN: Generative adversarial networks.
*e*: Training epochs of GAN.

Table 3: Selected variables in different scoring models

| Models | Original AutoScore | AutoScore-Imbalance | | | | | | | |
|---|---|---|---|---|---|---|---|---|---|
| | | SMOTE | US | DS | US + DS | SMOTE + DS | GAN ($e = 500$) | GAN ($e = 5000$) | GAN ($e = 5000$) + DS |
| $m$ [a] | 11 | 10 | 10 | 5 | 9 | 8 | 15 | 9 | 10 |
| Temperature (°C) | ● | ● | ● | ● | ● | | ● | | |
| Heart rate (beats/min) | ● | ● | ● | | ● | ● | ● | ● | ● |
| Age (years) | ● | ● | ● | ● | ● | ● | ● | ● | ● |
| Respiration rate (breaths/min) | ● | ● | ● | ● | ● | ● | ● | | |
| Systolic blood pressure (mm Hg) | ● | ● | ● | | ● | ● | ● | | |
| SpO$_2$ (%) | ● | ● | ● | | ● | ● | ● | | ● |
| White blood cells (thousand per microliter) | ● | ● | ● | | ● | | ● | | |
| Diastolic blood pressure (mm Hg) | ● | | | | | | ● | | |
| Platelet (thousand per microliter) | ● | ● | ● | | | | ● | | |
| Glucose (mg/dL) | ● | | | | | | | | |
| Sodium (mmol/L) | | | | | | | ● | ● | ● |
| Lactate (mmol/L) | ● | ● | ● | ● | ● | ● | ● | ● | ● |
| Mean arterial pressure (mm Hg) | | | | | | | ● | | |
| Potassium (mmol/L) | | | | | | | | | |
| Bicarbonate (mmol/L) | | | | | | | ● | ● | ● |
| Blood urea nitrogen (mg/dL) | | ● | ● | ● | ● | ● | ● | ● | ● |
| Hematocrit (%) | | | | | | | | | |
| Creatinine (μmol/L) | | | | | | | | ● | ● |
| Hemoglobin (g/dL) | | | | | | | | | |
| Chloride (mEq/L) | | | | | | | ● | ● | ● |
| Anion gap (mEq/L) | | | | | | ● | | ● | ● |

●: The variable is selected in this model

[a] Parameter $m$ is the number of variables included in the AutoScore model.

$e$: Training epochs of generative adversarial networks (GAN).

SMOTE: Synthetic minority over-sampling technique.

US: Up-sampling.

DS: Down-sampling.
GAN: Generative adversarial networks.
SpO$_2$: Peripheral capillary oxygen saturation.

Table 4. A summary of the nine-variable AutoScore-Imbalance (US+DS), five-variable AutoScore-Imbalance (DS), and eleven-variable original AutoScore scoring models for inpatient mortality prediction

| AutoScore-Imbalance | | | | Original AutoScore | |
|---|---|---|---|---|---|
| US+DS | | DS | | | |
| Variables and Interval [a] | Point | Variables and Interval | Point | Variables and Interval | Point |
| **Age (years)** | | | | | |
| <50 | 0 | <50 | 0 | <50 | 0 |
| 50-65 | 7 | 50-65 | 13 | 50-65 | 9 |
| 65-75 | 13 | 65-75 | 18 | 65-75 | 14 |
| ≥75 | 17 | ≥75 | 22 | ≥75 | 19 |
| **Lactate (mmol/L)** | | | | | |
| <1.7 | 6 | <1.7 | 13 | <1.7 | 7 |
| 1.7-1.8 | 2 | 1.7-1.8 | 0 | 1.7-1.8 | 4 |
| 1.8-2 | 0 | 1.8-2.3 | 5 | 1.8-1.95 | 0 |
| ≥2 | 13 | ≥2.3 | 24 | ≥1.95 | 12 |
| **Temperature (°C)** | | | | | |
| <36.5 | 4 | <36.5 | 5 | <36.5 | 6 |
| ≥36.5 | 0 | ≥36.5 | 0 | ≥36.5 | 0 |
| **Heart rate (beats/min)** | | | | | |
| <74 | 5 | | | <74 | 5 |
| 74-84 | 0 | | | 74-84 | 0 |
| 84-95 | 3 | | | 84-95 | 4 |
| ≥95 | 16 | | | ≥95 | 16 |
| **SpO$_2$ [b] (%)** | | | | | |
| <96.3 | 3 | | | <96.3 | 4 |
| 96.3-98.8 | 0 | | | 96.3-97.6 | 1 |
| ≥98.8 | 6 | | | 97.6-98.7 | 0 |
| | | | | ≥98.7 | 5 |
| **Systolic blood pressure (mm Hg)** | | | | | |
| <110 | 8 | | | <110 | 5 |
| 110-120 | 1 | | | 110-130 | 0 |
| 120-130 | 0 | | | ≥130 | 5 |
| ≥130 | 5 | | | | |

| White blood cells (thousand per microliter) | | | | | |
|---|---|---|---|---|---|
| <7.95 | 6 | | | <7.9 | 5 |
| 7.95-10.7 | 0 | | | 7.9-10.6 | 0 |
| ≥10.7 | 12 | | | 10.6-14 | 11 |
| | | | | ≥14 | 13 |
| **Respiration rate (breaths/min)** | | | | | |
| <16 | 2 | <16 | 2 | <16 | 3 |
| 16-18 | 0 | 16-18 | 0 | 16-18 | 0 |
| 18-21 | 3 | 18-21 | 4 | 18-20 | 5 |
| ≥21 | 7 | ≥21 | 16 | ≥20 | 6 |
| **Blood urea nitrogen (mg/dL)** | | | | | |
| <12.5 | 1 | <13.5 | 7 | | |
| 12.5-18 | 0 | 13.5-19 | 0 | | |
| 18-29 | 8 | 19-32.5 | 20 | | |
| ≥29 | 17 | ≥32.5 | 33 | | |
| **Platelet (thousand per microliter)** | | | | | |
| | | | | <160 | 7 |
| | | | | 160-210 | 4 |
| | | | | 210-280 | 0 |
| | | | | ≥280 | 3 |
| **Glucose (mg/dL)** | | | | | |
| | | | | <111 | 4 |
| | | | | 111-129 | 0 |
| | | | | 129-153 | 4 |
| | | | | ≥153 | 2 |
| **Diastolic blood pressure (mm Hg)** | | | | | |
| | | | | <71.1 | 8 |
| | | | | 71.1-77.2 | 6 |
| | | | | 77.2-85.2 | 0 |
| | | | | ≥85.2 | 5 |

[a] Interval (q1-q2) represents q1 ≤ x < q2.

[b] $SpO_2$: peripheral capillary oxygen saturation.

# Reference

1. Li, M. and G.B. Chapman, *Medical decision making.* The Wiley Encyclopedia of Health Psychology, 2020: p. 347-353.
2. Gultepe, E., et al., *From vital signs to clinical outcomes for patients with sepsis: a machine learning basis for a clinical decision support system.* Journal of the American Medical Informatics Association, 2014. **21**(2): p. 315-325.
3. Du, X., et al., *Predicting in-hospital mortality of patients with febrile neutropenia using machine learning models.* International journal of medical informatics, 2020. **139**: p. 104140.
4. Churpek, M.M., et al., *Derivation of a Cardiac Arrest Prediction Model Using Ward Vital Signs*. 2011, Am Heart Assoc.
5. Sullivan, L.M., J.M. Massaro, and R.B. D'Agostino Sr, *Presentation of multivariate data for clinical use: The Framingham Study risk score functions.* Statistics in medicine, 2004. **23**(10): p. 1631-1660.
6. Doshi-Velez, F. and B. Kim, *Towards a rigorous science of interpretable machine learning.* arXiv preprint arXiv:1702.08608, 2017.
7. Lundberg, S. and S.-I. Lee, *A unified approach to interpreting model predictions.* arXiv preprint arXiv:1705.07874, 2017.
8. Ribeiro, M.T., S. Singh, and C. Guestrin. *" Why should i trust you?" Explaining the predictions of any classifier*. in *Proceedings of the 22nd ACM SIGKDD international conference on knowledge discovery and data mining*. 2016.
9. Rudin, C., *Stop explaining black box machine learning models for high stakes decisions and use interpretable models instead.* Nature Machine Intelligence, 2019. **1**(5): p. 206-215.
10. Royal College of Physicians. National Early Warning, S., *Standardizing the assessment of acute-illness severity in the NHS.* National Early Warning Score (NEWS), 2012.
11. Subbe, C.P., et al., *Validation of a modified Early Warning Score in medical admissions.* Qjm, 2001. **94**(10): p. 521-526 %@ 1460-2393.
12. Churpek, M.M., et al., *Derivation of a cardiac arrest prediction model using ward vital signs.* Critical care medicine, 2012. **40**(7): p. 2102.
13. Leteurtre, S., et al., *Can generic paediatric mortality scores calculated 4 hours after admission be used as inclusion criteria for clinical trials?* Critical care, 2004. **8**(4): p. 1-9.
14. Hecht, H., et al., *Clinical indications for coronary artery calcium scoring in asymptomatic patients: expert consensus statement from the Society of Cardiovascular Computed Tomography.* Journal of cardiovascular computed tomography, 2017. **11**(2): p. 157-168.
15. Greving, J.P., et al., *Development of the PHASES score for prediction of risk of rupture of intracranial aneurysms: a pooled analysis of six prospective cohort studies.* The Lancet Neurology, 2014. **13**(1): p. 59-66.
16. Xie, F., et al., *AutoScore: A Machine Learning–Based Automatic Clinical Score Generator and Its Application to Mortality Prediction Using Electronic Health Records.* JMIR medical informatics, 2020. **8**(10): p. e21798.

17. Xie, F., et al., *Score for Emergency Risk Prediction (SERP): An Interpretable Machine Learning AutoScore-Derived Triage Tool for Predicting Mortality after Emergency Admissions.* medRxiv, 2021.
18. Ang, Y., et al., *AKI Risk Score (AKI-RiSc): Developing an Interpretable Clinical Score for Early Identification of Acute Kidney Injury for Patients Presenting to the Emergency Department.* medRxiv, 2021.
19. Menardi, G. and N. Torelli, *Training and assessing classification rules with imbalanced data.* Data mining and knowledge discovery, 2014. **28**(1): p. 92-122.
20. Malof, J.M., M.A. Mazurowski, and G.D. Tourassi, *The effect of class imbalance on case selection for case-based classifiers: an empirical study in the context of medical decision support.* Neural Networks, 2012. **25**: p. 141-145.
21. Larrazabal, A.J., et al., *Gender imbalance in medical imaging datasets produces biased classifiers for computer-aided diagnosis.* Proceedings of the National Academy of Sciences, 2020. **117**(23): p. 12592-12594.
22. Mazurowski, M.A., et al., *Training neural network classifiers for medical decision making: The effects of imbalanced datasets on classification performance.* Neural networks, 2008. **21**(2-3): p. 427-436.
23. Somasundaram, A. and S. Reddy, *Parallel and incremental credit card fraud detection model to handle concept drift and data imbalance.* Neural Computing and Applications, 2019. **31**(1): p. 3-14.
24. Al-Ajlan, A. and A. El Allali, *CNN-MGP: Convolutional neural networks for metagenomics gene prediction.* Interdisciplinary Sciences: Computational Life Sciences, 2019. **11**(4): p. 628-635.
25. Zeng, M., et al. *Effective prediction of three common diseases by combining SMOTE with Tomek links technique for imbalanced medical data.* in *2016 IEEE International Conference of Online Analysis and Computing Science (ICOACS).* 2016. IEEE.
26. Rahman, M.M. and D.N. Davis, *Addressing the class imbalance problem in medical datasets.* International Journal of Machine Learning and Computing, 2013. **3**(2): p. 224.
27. Khalilia, M., S. Chakraborty, and M. Popescu, *Predicting disease risks from highly imbalanced data using random forest.* BMC medical informatics and decision making, 2011. **11**(1): p. 1-13.
28. Li, D.-C., C.-W. Liu, and S.C. Hu, *A learning method for the class imbalance problem with medical data sets.* Computers in biology and medicine, 2010. **40**(5): p. 509-518.
29. Goodfellow, I.J., et al., *Generative adversarial networks.* arXiv preprint arXiv:1406.2661, 2014.
30. Chou, Y.-C., et al., *Deep-learning-based defective bean inspection with GAN-structured automated labeled data augmentation in coffee industry.* Applied Sciences, 2019. **9**(19): p. 4166.
31. Yansaneh, I.S., *Construction and use of sample weights.* Designing Household Surveys Samples: Practical Guidelines, 2003.
32. Breiman, L., *Random Forests.* Machine Learning, 2001. **45**(1): p. 5-32.
33. Rendon, E., et al., *Data sampling methods to deal with the big data multi-class imbalance problem.* Applied Sciences, 2020. **10**(4): p. 1276.

34. Longadge, R. and S. Dongre, *Class imbalance problem in data mining review.* arXiv preprint arXiv:1305.1707, 2013.
35. Chawla, N.V., et al., *SMOTE: synthetic minority over-sampling technique.* Journal of artificial intelligence research, 2002. **16**: p. 321-357.
36. Torgo, L., *Data mining with R: learning with case studies.* 2016: CRC press.
37. Xu, L., et al., *Modeling tabular data using conditional gan.* arXiv preprint arXiv:1907.00503, 2019.
38. Jiang, S. and X. Lu, *WeSamBE: A weight-sample-based method for background subtraction.* IEEE Transactions on Circuits and Systems for Video Technology, 2017. **28**(9): p. 2105-2115.
39. He, H. and E.A. Garcia, *Learning from imbalanced data.* IEEE Transactions on knowledge and data engineering, 2009. **21**(9): p. 1263-1284.
40. Leevy, J.L., et al., *A survey on addressing high-class imbalance in big data.* Journal of Big Data, 2018. **5**(1): p. 1-30.
41. Thabtah, F., et al., *Data imbalance in classification: Experimental evaluation.* Information Sciences, 2020. **513**: p. 429-441.
42. Brodersen, K.H., et al. *The balanced accuracy and its posterior distribution.* in *2010 20th international conference on pattern recognition.* 2010. IEEE.
43. Robin, X., et al., *pROC: an open-source package for R and S+ to analyze and compare ROC curves.* BMC bioinformatics, 2011. **12**(1): p. 1-8.
44. Efron, B., *Bootstrap methods: another look at the jackknife*, in *Breakthroughs in statistics*. 1992, Springer. p. 569-593.
45. Feng, X., et al., *Package 'AutoScore'.* R package version, 2021.
46. Sedgewick, A.J., et al., *Learning mixed graphical models with separate sparsity parameters and stability-based model selection.* BMC bioinformatics, 2016. **17**(5): p. 307-318.
47. Ferentinos, K.P., *Deep learning models for plant disease detection and diagnosis.* Computers and Electronics in Agriculture, 2018. **145**: p. 311-318.
48. Gruber, T., et al. *On deep learning-based channel decoding.* in *2017 51st Annual Conference on Information Sciences and Systems (CISS).* 2017. IEEE.
49. Lym, S., et al. *PruneTrain: fast neural network training by dynamic sparse model reconfiguration.* in *Proceedings of the International Conference for High Performance Computing, Networking, Storage and Analysis.* 2019.
50. Luo, J., C.-M. Vong, and P.-K. Wong, *Sparse Bayesian extreme learning machine for multi-classification.* IEEE Transactions on Neural Networks and Learning Systems, 2013. **25**(4): p. 836-843.
51. Meinshausen, N. and P. Bühlmann, *Stability selection.* Journal of the Royal Statistical Society: Series B (Statistical Methodology), 2010. **72**(4): p. 417-473.
52. Fu, G.H., L.Z. Yi, and J. Pan, *LASSO‐based false‐positive selection for class‐imbalanced data in metabolomics.* Journal of Chemometrics, 2019. **33**(10): p. e3177.
53. Profitlich, H.-J. and D. Sonntag, *Interactivity and Transparency in Medical Risk Assessment with Supersparse Linear Integer Models.* arXiv preprint arXiv:1911.12119, 2019.
54. Pak, C., T.T. Wang, and X.H. Su, *An empirical study on software defect prediction using over-sampling by SMOTE.* International Journal of Software

55. Weng, L., *From gan to wgan.* arXiv preprint arXiv:1904.08994, 2019.
56. Zhang, J.J., et al., *Alternatives to the Kaplan–Meier estimator of progression-free survival.* The International Journal of Biostatistics, 2021. **17**(1): p. 99-115 %@ 1557-4679.
57. Gordon, B.A., et al., *Spatial patterns of neuroimaging biomarker change in individuals from families with autosomal dominant Alzheimer's disease: a longitudinal study.* The Lancet Neurology, 2018. **17**(3): p. 241-250 %@ 1474-4422.
58. Xie, F., et al., *AutoScore-Survival: Developing interpretable machine learning-based time-to-event scores with right-censored survival data.* arXiv preprint arXiv:2106.06957, 2021.

Engineering and Knowledge Engineering, 2018. **28**(06): p. 811-830 %@ 0218-1940.